\documentclass[sigconf]{acmart}
\usepackage[utf8]{inputenc}

\usepackage{flushend}
\usepackage[a-1b]{pdfx}   
\usepackage{balance}

\usepackage{color}
\usepackage{tcolorbox}
\usepackage[labelfont=bf]{caption}
\usepackage{subcaption}
\usepackage{hyperref}
\usepackage{makecell}
\usepackage{fixltx2e}
\usepackage{rotating,multirow}
\usepackage{etoolbox,xspace}
\usepackage{algorithmicx}
\usepackage{algorithm}
\usepackage{fontawesome}
\usepackage{algpseudocode}
\usepackage{wrapfig}
\usepackage{scalerel} 
\usepackage{booktabs}
\usepackage{multirow}
\usepackage{siunitx}
\usepackage{amsmath}
\usepackage{colortbl}   
\usepackage{xcolor}     

\algtext*{EndWhile}
\algtext*{EndIf}
\algtext*{EndFunction}
\algtext*{EndFor}
\algrenewcommand\algorithmicrequire{\textbf{Input:}}
\algrenewcommand\algorithmicensure{\textbf{Output:}}

\usepackage{amsthm}

\usepackage{tabularx}
\usepackage{ragged2e}  
\usepackage{booktabs}  
\usepackage{textcomp}
\usepackage{url}
\usepackage{comment}
\usepackage{enumitem}
\usepackage[simplified]{pgf-umlcd}
\usepackage{arydshln}   
\usepackage{tikz}
\usetikzlibrary{shapes.geometric, arrows.meta, positioning, calc}

\newcolumntype{Y}{>{\RaggedRight\arraybackslash}X}

\newtheorem{definition}{Definition}
\newtheorem{theorem}{Theorem}

\newtcolorbox{defbox}{
  colback=orange!10,
  colframe=orange!20,
  arc=2mm, 
  fonttitle=\bfseries,
  boxrule=0mm,
  boxsep=1mm,
  left=0mm,
  right=0mm,
  top=0mm,
  bottom=0mm
}

\newcounter{example}
\renewcommand{\theexample}{\arabic{example}}

\newtcolorbox{examplebox}{
  colback=blue!10,
  colframe=blue!20,
  arc=2mm, 
  fonttitle=\bfseries,
  boxrule=0mm,
  boxsep=1mm,
  left=0mm,
  right=0mm,
  top=0mm,
  bottom=0mm
}

\newenvironment{example}{
  \refstepcounter{example}
  \begin{examplebox}{}
    \noindent {\bf Example \theexample:} 
  }
  {\end{examplebox}} 

\newcommand{\squishlist}{
 \begin{list}{$\bullet$}
  { \setlength{\itemsep}{0pt}
     \setlength{\parsep}{1pt}
     \setlength{\topsep}{1pt}
     \setlength{\partopsep}{0pt}
     \setlength{\leftmargin}{1em}
     \setlength{\labelwidth}{1em}
     \setlength{\labelsep}{0.5em} } }

\newcommand{\squishend}{
  \end{list}
}

\definecolor{low}{rgb}{1, 0.8, 0.8}   
\definecolor{mid}{rgb}{1, 1, 0.8}     
\definecolor{high}{rgb}{0.8, 1, 0.8}  
\definecolor{DarkGreen}{rgb}{0.0, 0.8, 0.0}
\newcommand{\uparrowgreen}{\textcolor{DarkGreen}{$\uparrow$}}
\newcommand{\downarrowred}{\textcolor{red}{$\downarrow$}}
\newcommand{\downarrowgreen}{\textcolor{DarkGreen}{$\downarrow$}}
\newcommand{\uparrowred}{\textcolor{red}{$\uparrow$}}

\definecolor{americanrose}{rgb}{1.0, 0.01, 0.24}
\definecolor{airforceblue}{rgb}{0.36, 0.54, 0.66}
\definecolor{ao(english)}{rgb}{0.0, 0.5, 0.0}
\definecolor{ao}{rgb}{0.0, 0.0, 1.0}

\newcommand{\eat}[1]{}

\usepackage{xspace}

\newcommand{\ee}{\mathbb{E}}
\newcommand{\eps}{\varepsilon}
\newcommand{\Elements}{\mathcal{I}}
\newcommand{\LLM}{\mathcal{L}}
\newcommand{\Exposure}{\mathcal{X}}
\newcommand{\Relevance}{Rel}
\newcommand{\Helper}{\mathcal{H}}

\title{Rank It, Then Ask It: Input Reranking for Maximizing the Performance of LLMs on Symmetric Tasks}

\author{Mohsen Dehghankar}
\affiliation{%
    \institution{ University of Illinois Chicago}
    \city{}
    \country{}
}
\email{mdehgh2@uic.edu}

\author{Abolfazl Asudeh}
\affiliation{%
    \institution{ University of Illinois Chicago}
    \city{}
    \country{}
}
\email{asudeh@uic.edu}

\copyrightyear{2025}
\acmYear{2025}
\setcopyright{rightsretained}
\acmConference[SIGMOD-Companion '25]{Companion of the 2025 International Conference on Management of Data}{}{}
\acmBooktitle{Companion of the 2025 International Conference on Management of Data (SIGMOD-Companion '25)}
\acmDOI{}
\acmISBN{}

\begin{document}

\begin{abstract}
Large language models (LLMs) have quickly emerged as practical and versatile tools that provide new solutions for a wide range of domains. In this paper, we consider the application of LLMs on symmetric tasks where a query is asked on an (unordered) bag of elements. Examples of such tasks include answering aggregate queries on a database table. In general, when the bag contains a large number of elements, LLMs tend to overlook some elements, leading to challenges in generating accurate responses to the query. LLMs receive their inputs as ordered sequences. However, in this problem, we leverage the fact that the symmetric input is not ordered, and reordering should not affect the LLM's response. 

Observing that LLMs are less likely to miss elements at certain positions of the input, we introduce the problem of LLM input reranking: to find a ranking of the input that maximizes the LLM's accuracy for the given query without making explicit assumptions about the query. Finding the optimal ranking requires identifying (i) the relevance of each input element for answering the query and (ii) the importance of each rank position for the LLM's attention. We develop algorithms for estimating these values efficiently utilizing a helper LLM. We conduct comprehensive experiments on different synthetic and real datasets to validate our proposal and to evaluate the effectiveness of our proposed algorithms. Our experiments confirm that our reranking approach improves the accuracy of the LLMs on symmetric tasks by up to $99\%$ proximity to the optimum upper bound. 
\end{abstract}

\keywords{LLMs for Data Management; Ranking;}
\settopmatter{printfolios=true}

\maketitle

\section{Introduction}\label{sec:intro}

Large Language Models (LLMs) have rapidly become invaluable tools, expanding their impact far beyond the realm of natural language processing. Tasks that have long been studied across various areas of computer science are now finding alternative solutions with LLMs.

In particular, LLMs can be used for symmetric~\cite{gao2023double} tasks where a query is issued on an input that takes the form of a bag (or multi-set) of elements. Querying a database relation is an example of this type of task.
To further clarify this, let us consider the following toy example.

\begin{examplebox}
\begin{example} \label{ex:db}
{\bf (Course Registration Database)}
Consider the textbook example of the course registration database with various tables such as {\tt Student}, {\tt Professor}, {\tt Department}, {\tt CourseRegistration}, etc.
Each row of table {\tt CourseRegistration} contains a {\tt CourseID} registration for a {\tt StudentID} at a specific {\tt Semester}:

\begin{center}
\begin{tabular}{|c|c|c|}
    \hline
    {\tt\bf  StudentID} & {\tt\bf  CourseID} & {\tt\bf Semester}  \\ \hline
    10023415 & CS480 & Fall24 \\ \hline
    10042652 & CS401 & Spring23 \\ \hline
    ...&...&... \\ \hline
\end{tabular}
\end{center}
Suppose one is interested in finding the number of students registered for the Database Systems (CS480) course in Fall 2024.
They can specify their query in the form of a prompt\footnotemark
{\tt ``Count the number of rows where CourseID is CS480 and Semester is Fall24''}, and pass it alongside the table {\tt CourseRegistration} to an LLM to find the answer.
\end{example}
\end{examplebox} \footnotetext{A prompt is a textual instruction for the LLM.}

\begin{figure}[!tbh]
    \centering
    \includegraphics[width=0.9\linewidth]{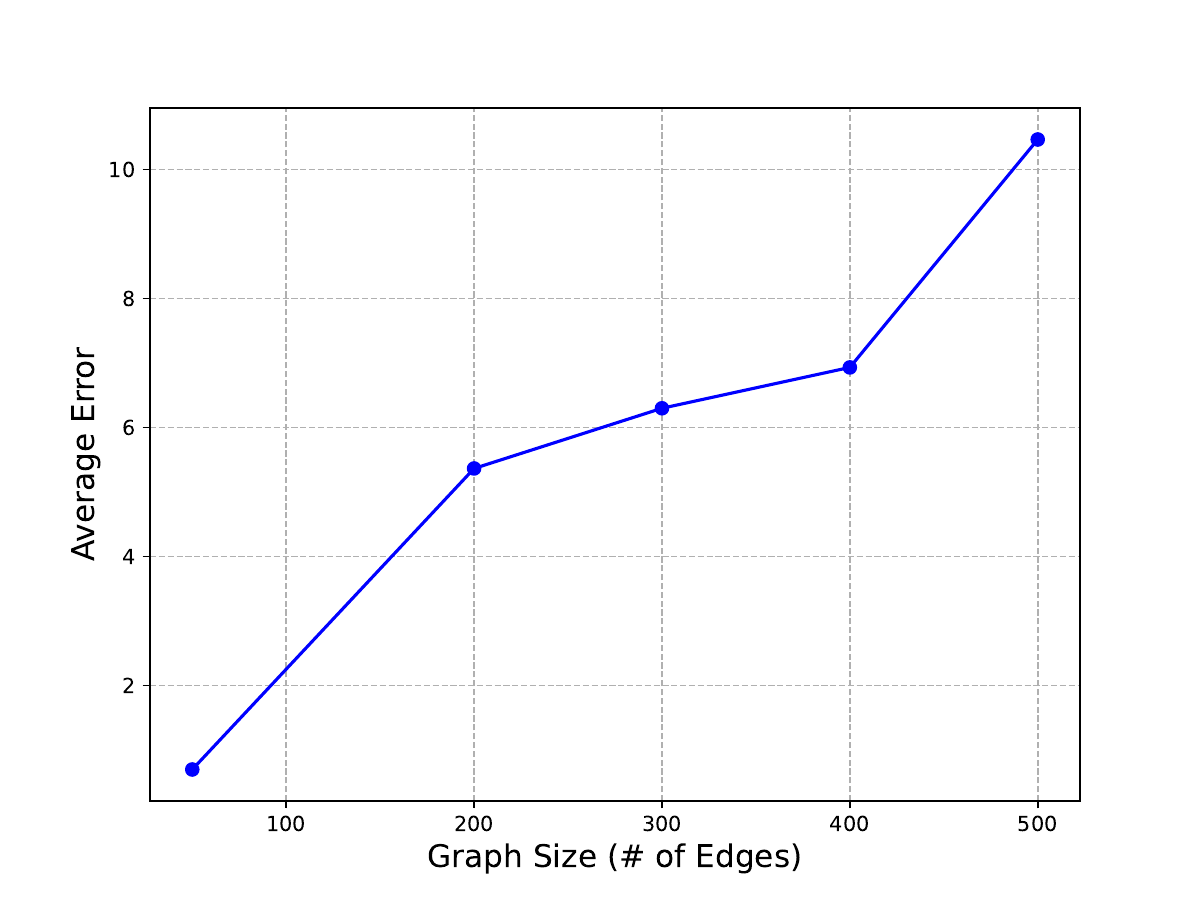}
    \caption{Illustrating the average error of GPT-3.5 Turbo on Graph Degree Task based on different input sizes. The error is the absolute difference between the real degree of a node (less than 20 in this case) and the reported degree by the LLM.}
    \label{fig:error_demo}
\end{figure}

Examples of symmetric tasks are not limited to databases. For example, passing the edges of a graph as the input elements, one can ask the LLM for the degree of a specific node (see Example~\ref{ex:1}). We refer to this as the Graph Degree Task.

LLMs follow a sequential randomized process to generate their outputs. As a result, their inputs are received and processed as ordered sequences. 
In particular, when the input is lengthy, LLMs are known to pay ``less attention'' to certain regions of the input, hence, struggling with retaining information from those regions. This leads to a performance degradation over extended inputs ~\cite{liu2024lost}. 
For example, Fig ~\ref{fig:error_demo} shows the significant error of GPT-3.5 Turbo for Graph Degree Task when the graph size is more than 200.

On the other hand, the input of a symmetric task is an unordered bag of elements. As a result, the input elements can be freely reordered.
This gives us the opportunity to {\em rerank the input elements} before passing it to the LLM for maximizing its query-answering accuracy -- which is the research focus of this paper.
Specifically, we introduce a reranking mechanism that (a) learns the so-called ``exposure'' of each rank position for an LLM during the preprocessing time. Then, at the query time, it (b) assesses the relevance of each input element to the given natural language query and (c) generates a reranking of the input that puts the relevant elements in high-exposure positions.


\subsection{Summary of Contributions}
In this work, {\em we introduce the problem of LLM input reranking} to find an ordering of the input that maximizes the accuracy of an LLM for symmetric tasks. To the best of our knowledge, we are the first to introduce and study this problem.

We propose a two-stage solution for the problem.
First, we identify the exposures of the LLM to detect patterns of ``forgetting'' when processing large inputs. Second, we rank the input elements 
by estimating their relevance scores in relation to a query $q$, without requiring explicit knowledge of the task or the query itself. Our method enhances the effectiveness of LLMs in handling large and complex inputs in symmetric problem scenarios.

We present extensive experimental evaluations across two distinct categories of tasks. First, in a Graph Degree Task, we demonstrate that by utilizing lightweight versions of open-source large language models, such as Llama 3.1, Qwen 2, and DeepSeek-Coder, we are able to estimate relevance scores and rank input elements with rank utility values approaching those of optimal solutions (see Table~\ref{tab:rank-util}). 
Second, leveraging the obtained ranking information, we introduce a reranking approach that significantly enhances the accuracy of LLM outputs. {\em Our method improves performance by up to {\bf 99\%} proximity to the optimum solution}, as demonstrated in Table ~\ref{tab:errs-all}.

We also identified a notable distinction in the token retention patterns of two widely used commercial LLMs. 
For instance, we observe that
GPT-3.5 Turbo demonstrates a stronger tendency to retain tokens positioned at the beginning of the prompt, whereas in GPT-4o Mini, tokens located in the middle are more likely to be remembered by the model (Fig. ~\ref{fig:exposures}).

Additionally, we conducted experiments on other tasks, such as database query answering with real-world datasets, and observed comparable improvements in the output of the LLMs after reranking the input.

\subsection{Paper Organization}
The rest of the paper is organized as follows: first, in Section~\ref{sec:prelim} we introduce the necessary notations and formalize our input reranking problem.
Next, we present our algorithm for estimating the relevance of the input elements to the given query, in Section~\ref{sec:rel}.
In Section~\ref{sec:exposure}, we discuss our approach for learning the exposure values for an LLM for each rank position.
Our experimental evaluations are provided in Section~\ref{sec:exp}, followed by the related work, discussions, and the conclusion in Sections~\ref{sec:related}, \ref{sec:discussion}, and \ref{sec:conclusion}.

\tikzset{
    user/.style = {rectangle, draw, rounded corners, fill=blue!15, minimum height=3cm, minimum width=4cm, text width=4cm, align=left, font=\small},
    user2/.style = {rectangle, draw, rounded corners, fill=blue!15, minimum height=1cm, minimum width=4cm, text width=4cm, align=left, font=\small},
    llm/.style = {rectangle, draw, rounded corners, fill=green!15, minimum height=1cm, minimum width=4cm, text width=4cm, align=left, font=\small},
}

\section{Preliminaries}\label{sec:prelim}
In this section, we formally define the problem and the specific notations that we use in the following sections.


\subsection{Query Model}

We assume that each task $T$ is a pair $(\Elements, q)$. Where $\Elements$ is a long list of symmetric (bag of) elements  $\{e_1, e_2, ..., e_n\}$ and $q$ is a query in natural language about $\Elements$.

For example, for the Graph Degree Task, $\Elements$ is the list of edges with any order and $q$ is a natural language question about the graph, like "What is the degree of node 10?". In other example, for the Database Query Task, $\Elements$ is the list of rows of a database table and $q$ is a SQL query or a query in natural language about the table, for example, "How many records have the attribute $A_1 > 100$?". We can assume that $\Elements$ and $q$ are given to the LLM as two different prompts in the same context. That is, we first provide the list to the LLM, then we ask about the query $q$. More details on the implementation is discussed in section ~\ref{sec:exp}.

\subsection{LLM Model}
We assume that we have API access to a black-box LLM $\LLM$. For a task $T = (\Elements, q)$ as defined in the previous section, $\LLM(\Elements, q)$ is the response of the LLM. The output error is defined as: 

\[
    \eps_{\LLM}(\Elements, q) = \Delta[\LLM(\Elements, q), \mathcal{O}(\Elements, q)]
\]

Where $\mathcal{O}(\Elements, q)$ is defined as the correct output of the task, and $\Delta$ is a function that measures the distance between the LLM results and the correct output. For example, for the Graph Degree Task, $\Delta$ is the absolute difference between the reported degree and the real degree of the node.

\subsection{Problem Definition}\label{sec:problem-def}
Given a task \( (\Elements, q) \) and a large language model \( \LLM \), our objective is to rerank the elements in \( \Elements \) with respect to the query \( q \) to minimize the expected error, denoted as \( \mathbb{E}(\eps_{\LLM}(\Elements, q)) \). Specifically, we seek to identify the optimal reranking function \( \pi^*: \{1, 2, \dots, n\} \rightarrow \{1, 2, \dots, n\} \) from the set of all possible rankings \( \Pi \). This function, \( \pi^* \), rearranges each element \( e_{\pi^*(i)} \) to a new position \( i \) in such a way that the response of the LLM exhibits an improvement in terms of reduced expected error. We denote the reordered list of elements as \( \Elements_{\pi^*} \).

\begin{align}\label{eq:optimization}
 \pi^* = \arg\min_{\pi \in \Pi} \mathbb{E}[\eps_{\LLM}(\Elements_{\pi}, q)]
\end{align}

Our approach is task (and query) agnostic. In other words, we find the function $\pi^*$ without using any {\bf explicit knowledge} about the query or the task. 

\subsection{Solution Overview}\label{sec:sol}

We define a {\em utility} function on different orderings of $\Elements$. Based on the objective function in our Problem Definition (Section~\ref{sec:problem-def}), the {\em utility} of a reranking function $\pi$ should capture the expected error $\mathbb{E}\left[\eps_{\LLM}(\Elements_{\pi}, q)\right]$.

Let us define a function $\Relevance_q: \Elements \rightarrow [0, 1]$ that captures the relevance of each element $e_i \in \Elements$ to the query $q$. That is, $\Relevance_q(e_i)$ is the relevance of $e_i$ to the query $q$. 

Also, let $\Exposure_{\LLM}(i)$ denote the ``exposure'' of the position $i$ in a ranked input to the LLM $\LLM$, i.e., the likelihood that the LLM will not miss an element in position $i$. 
Then, the expected {\em utility} of a ranking $\pi$ of the input $\Elements$ is calculated as~\cite{singh2018fairness},

\begin{align}\label{eq:rerank2}
    \mathbb{E}\left[utility(\pi|q)\right] = \sum^{|\Elements|}_{i=1} \mathbb{E}\left[\Exposure_{\LLM}(i)\right] \cdot \mathbb{E}\left[Rel_q(e_{\pi(i)})\right]
\end{align}

\begin{examplebox}
\begin{example}\label{ex:1} 
      Consider the graph $G$, with 6 vertices and the following edges:

\begin{center}
\begin{tabular}{c|c|c|c|c|c|c|c|c|c}
    $e_1$ & $e_2$ & $e_3$ & $e_4$ & $e_5$ & $e_6$ & $e_7$ & $e_8$ & $e_9$ & $e_{10}$  \\ \hline
    1&2&1&3&1&2&3&3&5&2 \\ \hline
    2&4&4&4&3&5&5&6&6&6 \\
\end{tabular}
\end{center}
Also, let the exposure function be $\Exposure_{\LLM}(i)=\frac{1}{i}$.
Now let the query $q$ be {\tt ``compute the degree of $v_1$''}.
For edges incident to $v_1$, $\Relevance_q(e_i)=1$, while for the others the relevance is 0.
As a result, the utility of the ranking $\pi = \{1,2,\cdots,10\}$ is $utility(\pi|q) = 1 + \frac{1}{3} + \frac{1}{5} \simeq 1.53$. Note that the ranking with maximum utility put $e_1$, $e_3$, and $e_5$ at the beginning of the list, and has the utility of $1+\frac{1}{2}+\frac{1}{3}\simeq 1.83$.
\end{example}
\end{examplebox}

The first step towards computing the utility of a ranking is to specify the exposure function $\Exposure$. We note that {\em the exposure values are LLM-specific}, i.e., the exposure of a position in the ranking may vary based on the LLM at hand and the length of each element in the input $\Elements$. 
Therefore, during the preprocessing time, we need to estimate the exposure values for a given LLM. To do so, 
in Section~\ref{sec:exposure}, we develop an approach based on sample tasks for which the query $q$, the input $\mathcal{I}$, the correct output $\mathcal{O}$, and the $\Relevance_q(e_i), \forall e_i\in\Elements$ is known ahead of time.

After specifying the exposure values during the preprocessing time, we need to estimate the relevance values at the query time.
Recall that $q$ can be any natural language query on the input set $\Elements$.
Given a task $T=(\Elements,q)$, in Section~\ref{sec:rel}, we present our approach for estimating the relevance function $\Relevance_q$. 
\section{Estimating the Relevance}\label{sec:rel}

Each element \( e_i \in \Elements \) is associated with an implicit relevance score that quantifies its degree of relevance to the query \( q \). This relevance score is denoted as \( \Relevance_q(e_i) \). 
For instance, in Example~\ref{ex:1}, the edges adjacent to the node $v$ are more relevant to the query ``compute the degree of node $v$''. 

However, since the task $T$ is not specified beforehand, and $q$ can be any query on $\Elements$, computing the relevance values is challenging.
Therefore, in this section, we tackle this issue by
studying the sub-problem of estimating the
relevance score for each element $e_i\in \Elements$.

We utilize a helper LLM $\Helper$ for estimating the relevance values.
{\em $\Helper$ is a small open-source LLM that is relatively cheap to run and deploy}. In our experiments, we compare different models as $\Helper$, like Gemma2 (9b)~\cite{team2024gemma}, Llama3.1 (8b)~\cite{dubey2024llama}, Qwen2 (7b)~\cite{yang2024qwen2}, DeepSeek-Coder-v2 (16b)~\cite{zhu2024deepseek}, and Mistral (7b)~\cite{jiang2023mistral}.

In the rest of this section, we explain two methods to estimate the relevance scores. First, we will discuss a warm-up baseline, in which the input list \( \Elements \) is partitioned into smaller subsets to split relevant and non-relevant elements. 
Next, we present our estimation approach based on modeling the problem as a bipartite graph.


\subsection{Warm-up: Partitioning the Input}\label{sec:warm-up} 

Let us consider Example~\ref{ex:1} once again.
To find the relevant edges to the given query, one can partition the input elements $\Elements$ into smaller subsets and ask the helper LLM to find relevant edges in each subset.

Following this idea, 
the warm-up algorithm first partitions the input list $\Elements$ into $m$ smaller subsets of size $l=\lceil\frac{n}{m}\rceil$, i.e.,
$\{P_1=\{e_1,\cdots,e_l\}, P_2=\{e_{l+1},\cdots,e_{2l}\},\cdots, P_m= \{e_{(m-1)l+1},\cdots,e_{ml}\}\}$.
Next, for each partition $P_i$, the algorithm asks the helper LLM $\Helper$
``{\tt what elements in [$P_i$] are more relevant for answering the query $q$}?''. 
It then assigns the relevance score $1$ to the returned elements and $0$ to others. See Algorithm ~\ref{alg:psm} for the more details.

\begin{algorithm}[!tb]
    \caption{Warm-up} \label{alg:psm}
    \begin{algorithmic}[1] \small
    \Require{The list of elements $\Elements$, The query $q$, The helper LLM $\Helper$, Number of partitions $m$}
    \Ensure{The list of relevance scores $\{\Relevance_q(e_i) \;|\; e_i \in \Elements$\}}
    \Function{PSM}{$\Elements$, $q$, $\Helper$, $m$}
        \State Partition $\Elements$ into $m$ chunks
        \State $P \gets $ All chunks
        \For{$P_i \in P$}
            \State $R \gets \mathcal{H}(P_i, q)$\Comment{Ask helper to give the relevant elements $R$}
            \State $\Relevance_q(e_i) \gets 1$ for all $e_i \in R$
            \State $\Relevance_q(e_i) \gets 0$ for all $e_i \in P_i \setminus R$
        \EndFor
        \State \bf{Return} $\{\Relevance_q(e_i) \;|\; e_i \in \Elements\}$
    \EndFunction
    \Statex
    \end{algorithmic}
\end{algorithm}

\subsection{Modeling of the Relevance Estimation as Bipartite Graph} \label{sec:bipartite}

The warm-up algorithm, while providing a baseline for estimating the relevance values, suffers from multiple drawbacks.
First, the scope of the relevance values generated by the warm-up algorithm is limited to binary. 
Second, given the randomized nature of the LLMs, the output generated by the zero-shot process of the warm-up algorithm may not be reliable; particularly, given that we assume we have no prior knowledge about the task and the query at hand. 
Third, depending on the composition of a partition, the output may miss relevant elements or return less relevant ones.
In other words, the helper LLM may over/under-estimate the relevance scores in each partition.
As a result, changing the partition an element belongs to, may impact its relevance-value estimation.

To address the first issue, one can ask the helper to directly estimate the relevance value of each element in each partition, but this would result in high variance and sometimes inconsistent scores.
Alternatively, addressing the first and second issues is possible by collecting multiple responses for each partition -- instead of relying on a single evaluation. The relevance value of each element is then the average of each individual estimation.
This, however, does not resolve the third issue; hence the estimations may remain inaccurate.
Addressing the third issue is possible by shuffling the input before each partitioning, but this may require a large number of evaluations to generate unbiased and accurate estimations for each element -- making the relevance estimation process costly. 

Instead, our goal is to obtain accurate estimations with {\em a small number of evaluations per element}.


Let us now make a connection to the peer-reviewing process~\cite{liu2022integrating}, in which a small number of reviewers review each paper, while various reviewers may generally provide higher/lower scores for the papers they review.
Similarly, we can view each element as a paper and each relevance-value estimation for the elements in a partition as the scores provided by a reviewer.


Inspired by this connection, we devise a similar process for relevance-value estimation. 
Specifically, we randomly shuffle the input list $\Elements$ a total of $\sigma$ times (for a small value of $\sigma$) to get the shuffled lists $\{\Elements_1, \Elements_2, ..., \Elements_{\sigma}\}$.
Subsequently, we partition each shuffled list into $m$ equal-size subsets, as in the warm-up algorithm.
This ensures that each evaluation of an element will be with a different set of elements, minimizing the partition composition impact on the final evaluations.
Let $P_{i, k}$ denote the $k^{\text{th}}$ partition of $\Elements_i$ where $i \leq \sigma$ and $k \leq m$. We ask the helper LLM $\Helper$ to give us a categorized score (e.g., from one to five) for each element inside each of these $P_{i, k}$ partitions. We index the score-evaluations  to all partitions as $\{\mathcal{E}_{1}, \mathcal{E}_{2},\cdots, \mathcal{E}_{\sigma m}\}$ 
(the evaluations obtained for $P_{i,k}$ is $\mathcal{E}_{(i-1)m+k}$).
Each evaluation can be considered a potentially biased (i.e., over/under-estimated) set of scores assigned to a set of elements. 

For an element $e_i$, let $S_i=Rel_q(e_i)$.
Assume we could obtain a collection of $\sigma$ unbiased evaluations $\{w^{o}_{i,1},w^{o}_{i,2},\cdots,w^{o}_{i,\sigma}\}$ for $e_i$. Then, each $w^{o}_{i,j}$ would be viewed as a random variable taken from a distribution with mean $S_i$, i.e., $\ee\left[w^{o}_{i,j} \right] = S_i$.
Then \(\bar{S}_i=\frac{1}{m}\sum_{j=1}^\sigma w^{o}_{i,j}\) would be an unbiased estimation for $S_i$.

Let $w_{i,j}$ be the evaluation score of $e_i$ in $\mathcal{E}_j$.
We assume that each evaluation $\mathcal{E}_j$ equally over/under estimates the evaluation scores for all elements with an unknown but constant coefficient $\beta_j$. That is, for all $e_i$ evaluated in $\mathcal{E}_j$, 
\( w^o_{i,j} = \frac{1}{\beta_j} w_{i,j}\).
Had we known the bias values $\{\beta_1,\cdots, \beta_j\}$, we could estimate the relevance score $S_i$ of the element $e_i$ as
\(
    \bar{S}_i = \frac{1}{\sigma}\sum_{e_i\in\mathcal{E}_j} \frac{1}{\beta_j} w_{i,j}
\).

In order to find the bias coefficients for each of the evaluations, we build a bipartite graph $\mathcal{G}$, which we call the evaluation graph. 

\begin{definition}[Bipartite Evaluation Graph]\label{def:bipartite}
    Consider the weighted bipartite graph $\mathcal{G}(U, V, E)$. $U$ contains $n$ nodes, each representing an element $e_i\in\Elements$. $V$ is a set of $\sigma \cdot m$ nodes, representing the evaluation outputs $\mathcal{E}_1,\cdots \mathcal{E}_{\sigma m}$. 
    An edge $(u_i,v_j)$ belongs to $E$ iff $\mathcal{E}_j$ contains the element $e_i$, with its weight being equal to $w_{i,j}$ -- the evaluation score $\mathcal{E}_j$ provides for $e_i$.
\end{definition}

\begin{figure}[!tb]
\centering
\begin{tikzpicture}

    \node[draw, circle, minimum size=0.9cm, fill=red!30] (U1) at (0, 3) {\tiny $\bar{S}_1$};
    \node[draw, circle, minimum size=0.9cm, fill=red!30] (U2) at (0, 2) {\tiny $\bar{S}_2$};
    \node[draw, circle, minimum size=0.9cm, fill=red!30] (U3) at (0, 1) {\tiny $\bar{S}_{3}$};
    \node at (0, 0) {$\vdots$}; 
    \node[draw, circle, minimum size=0.9cm, fill=red!30] (Un) at (0, -1) {\tiny $\bar{S}_{n}$}; 

    \node[draw, circle, minimum size=0.9cm, fill=blue!30] (V1) at (3, 3) {\tiny $\beta_1$};
    \node[draw, circle, minimum size=0.9cm, fill=blue!30] (V2) at (3, 2) {\tiny $\beta_2$};
    \node[draw, circle, minimum size=0.9cm, fill=blue!30] (V3) at (3, 1) {\tiny $\beta_{3}$};
    \node at (3, 0) {$\vdots$}; 
    \node[draw, circle, minimum size=0.9cm, fill=blue!30] (Vn) at (3, -1) {\tiny $\beta_{\sigma m}$}; 

    \draw (U1) -- (V1) node[midway, above] {$w_{1, 1}$};;
    \draw (U1) -- (V2) node[midway, above] {$w_{1, 2}$};
    \draw (U2) -- (V2);
    \draw (U2) -- (Vn);
    \draw (U3) -- (V1);
    \draw (U3) -- (V3);
    \draw (Un) -- (V3);
    \draw (Un) -- (Vn);

    \node at (-1.5, 1.5) {$U$};
    \node at (4.5, 1.5) {$V$};

\end{tikzpicture}
\caption{An example of the bipartite representation of evaluations. Each node in $U$ represents an element in $\Elements$ and its final score. Each node in $V$ represents one evaluation done on a partition $P_{i, k}$ from a shuffled list $E_i$ and the associated bias with that. The weights on the edges are the scores assigned by the helper LLM $\Helper$ to elements.}
\label{fig:bipartite_graph}
\end{figure}
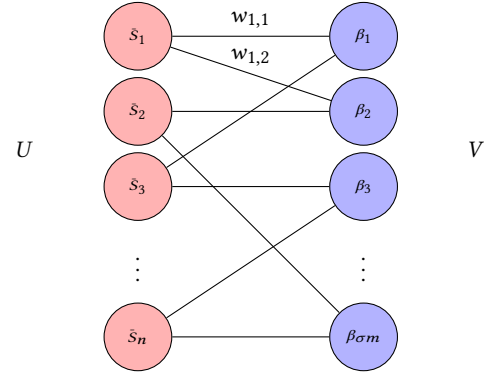

In the bipartite evaluation graph, the degree of each node $u_i\in U$ is $deg(u_i)=\sigma$, while the degree of each node $v_j\in V$ is $deg(v_j)=\lceil\frac{n}{m}\rceil$ (size of each partition). An illustration of the graph $\mathcal{G}$ is shown in Figure \ref{fig:bipartite_graph}.

Let us associate each node $u_i\in U$ with the weight $\bar{S}_i$ and each node $v_j\in V$ with the weight $\beta_j$.
Then, the following equations hold:

\begin{align}\label{eq:bipartite}
    \nonumber
    \bar{S}_i = \frac{1}{\sigma}\sum_{(u_i,v_j)\in E} \frac{w_{i,j}}{\beta_j}, ~& ~~\forall u_i\in U \\
    \beta_j = \frac{1}{\lceil\frac{n}{m}\rceil} \sum_{(u_i,v_j)\in E} \frac{w_{i, j}}{\bar{S}_i}, ~& ~~\forall v_j\in V
\end{align}


%

Therefore, to estimate the bias coefficients, we develop an iterative numeric process that at every iteration updates the node weights $\beta_j$ (resp., $\bar{S}_i$) based on its current estimates of the weights $\bar{S}_i$ (resp., $\beta_j$) of the nodes connected to it.


The process starts by initializing all $\beta_j$ values with 1 ($\beta_j^{(0)} = 1$).
It then alternates between updating $\bar{S}_i$ values based on $\beta_j$ values or we updating $\beta_j$ using $\bar{S}_i$ values:


\begin{align} \label{eq:update-s-beta}
    \nonumber
    \bar{S}_i^{(T)} = \frac{1}{\sigma} \sum_{(u_i,v_j) \in E} \frac{w_{i, j}}{\beta_j^{(T - 1)}}, ~& ~~\forall u_i\in U\\
    \beta_j^{(T + 1)} = \frac{1}{\lceil\frac{n}{m}\rceil} \sum_{(u_i,v_j) \in E} \frac{w_{i, j}}{\bar{S}_i^{(T)}}, ~& ~~\forall u_i\in U
\end{align}

Where $\bar{S}_i^{(T)}$ (resp., $\beta_j^{(T)}$) is the estimated relevance score (resp. bias coefficient) at time step $T$. A pseudo-code of this method is presented in Algorithm~\ref{alg:bem}.

\begin{theorem}
   The process described in Equation~\ref{eq:update-s-beta} would eventually converge.
\end{theorem}

{\bf Proof: } 
Define the matrix $W = (w_{i, j})$. Let $\bar{S} = (\bar{S}_i)$ be a vector of size $n$ and $\beta = (\beta_j)$ be a vector of size $\sigma \cdot m$. For a vector $v = (v_i)$ of size $n$, let $\frac{1}{v}$ denote the new vector $(\frac{1}{v_1}, \frac{1}{v_2}, \cdots, \frac{1}{v_n})$ and $v[k]$ denote the $k^{th}$ element of a vector $v$. 

Let $A \circ v$ denote the Hadamard product of an $n \times m$ matrix $A$ to vector $v$ of size $m$. The result of this production is a matrix $B$ of size $n \times m$ where $B(i, j) = A(i, j) \cdot v[j]$. We also use $\mathbf{1}_n$ to denote a vector of size $n$ with all elements equal to $1$.

We can rewrite Equation~\ref{eq:update-s-beta} using the new notations:

\begin{align}
    &\bar{S}^{(T)} = \frac{1}{\sigma} \cdot W \cdot \frac{1}{\beta^{(T - 1)}}\nonumber\\
    &\beta^{(T + 1)} = \frac{1}{\lceil \frac{n}{m} \rceil} \cdot W^{\top} \cdot \frac{1}{\bar{S}^{(T)}}\nonumber
\end{align}

Let us change the vector products to Hadamard products:

\begin{align}
    &\bar{S}^{(T)} = \frac{1}{\sigma} \cdot \left(W \circ \frac{1}{\beta^{(T - 1)}}\right) \cdot \mathbf{1}_{\sigma \cdot m}\label{eq:rewrite-1}\\
    &\beta^{(T + 1)} = \frac{1}{\lceil n / m \rceil} \cdot \left(W^{\top} \circ \frac{1}{\bar{S}^{(T)}} \right) \cdot \mathbf{1}_{n}\label{eq:rewrite-2}
\end{align}

We can now define a sequence of matrices $W_t$ according to the above update process:

\[
W_t = 
    \begin{cases} 
    W \circ \frac{1}{\beta^{(t / 2)}} & \text{if } t \; \text{is even}, \\
    \left(W^{\top} \circ \frac{1}{\bar{S}^{((t + 1) / 2)}}\right)^{\top} & \text{if } t \; \text{is odd}.
    \end{cases}
\]

For example, $W_0 = W \cdot \frac{1}{\beta^{(0)}} = W$ because $\beta^{(0)} = \mathbf{1}_{\sigma \cdot m}$. However, $W_1$ is a column-scaled of $W$, so the sum of each column is now equal to $\sigma$. Subsequently, $W_2$ is a row-scaled of $W_1$, so that the sum of each row is equal to $\lceil n / m \rceil$. This process continues similarly. At each iteration, we either rescale the rows or columns of $W_t$ to get the matrix $W_{t + 1}$. As a result, we can reduce this problem to the following:

Given a matrix $W$ of size $n \times (\sigma m)$, at each iteration, we are alternatively scaling the rows to have a sum of $\sigma$ or scaling the columns to have a sum of $\lceil n / m \rceil$. We would like to know the convergence condition for this process.

This is the same process as Sinkhorn's algorithm~\cite{sinkhorn1967concerning,sinkhorn1967diagonal} to find a doubly stochastic matrix starting from a positive matrix $W$. If $W$ is a non-negative matrix with at least one positive diagonal (refer to~\cite{sinkhorn1967concerning}), then this process would converge to a doubly stochastic matrix of $W$ such that in this matrix ($W_{\infty}$), the sum of each row is equal to $\sigma$ and the sum of each column is equal to $\lceil n / m \rceil$. We skip most of the details to the references. 

Now, we can rewrite the equations \ref{eq:rewrite-1} and~\ref{eq:rewrite-2} as:

\begin{align}
    \bar{S}^{(\infty)} = \frac{1}{\sigma} \cdot W_{\infty} \cdot \mathbf{1}_{\sigma \cdot m}\\
    \beta^{(\infty)} = \frac{1}{\lceil n / m \rceil} \cdot W_{\infty} \cdot \mathbf{1}_n
\end{align}. $\square$

\begin{algorithm}[!tb]
    \caption{Bipartite} \label{alg:bem}
    \begin{algorithmic}[1] \small
    \Require{The list of elements $\Elements$, The query $q$, The helper LLM $\Helper$, Number of partitions $m$, Number of shuffles $\sigma$}
    \Ensure{The list of estimated relevance scores $\{\Relevance_q(e_i) \;|\; e_i \in \Elements$\}}
    \Function{BEM}{$\Elements$, $q$, $\Helper$, $m$, $\sigma$}
        \State $G \gets $ Initialize the bipartite graph
        \For{$1 \leq i \leq \sigma$}
            \State $\Elements_i \gets $ Random shuffle $\Elements$
            \State $P_i \gets $ Partition $\Elements_i$ into $m$ chunks
            \For{$P_{i, k} \in P_i$}
                \State $\mathcal{E} \gets \Helper(P_{i, k})$\Comment{Ask helper to score this chunk}
                \State Update ${G}$ \Comment{Add edges $w_{i, j}$ based on observed scores.}
            \EndFor
        \EndFor
        \State Initialize $\beta_j$ values inside ${G}$
        \While{didn't converge}
            \State Update $\bar{S}_i$ and $\beta_j$ values based on Equations~\ref{eq:update-s-beta}
        \EndWhile
        \State {\bf Return} $\{\bar{S}_1, \bar{S}_2,\cdots, \bar{S}_n \}$ 
    \EndFunction
    \Statex
    \end{algorithmic}
\end{algorithm}

The above process requires in total $(\sigma m)$ evaluations, i.e., API calls to the helper LLM $\Helper$\footnote{Note that $m$ is not necessarily the same as $m$ defined in the previous method. Refer to the experiments Section~\ref{sec:exp} for more details on comparing these two methods.}.
\section{Pre-processing: Exposure Discovery}\label{sec:exposure}
So far, we studied the relevance estimation of the input elements to a given query.
The remaining information for the LLM-input reranking based on Equation~\ref{eq:rerank2} is to compute the exposure values for different rank positions.
The exposure values are LLM-dependent, and hence, we compute those during the preprocessing phase for each large language model $\LLM$.

The exposure value of the position $i$ in the ranking, i.e., $\Exposure_{\LLM}(i)$, represents the likelihood that the model misses each token in the input prompt as a function of its position $i$.
Throughout this section, $i$ will be used to indicate the position of a {\bf token} within the input prompt to $\LLM$. In the end, we shall explore how this information can be utilized for the general list $\Elements$, where each element $e_i \in \Elements$ is not necessarily a token.

In order to estimate the exposure values, we consider a sample set of predefined tasks, each consisting a query $q$ and the input elements $\Elements = [t_1, t_2, \dots, t_n]$ (consisting of $n$ tokens arranged in sequential order).
The task samples can be viewed as the training data we use to learn the exposure values.
Note that for each of the tasks, we already know the ground-truth relevance values $\Relevance_q(.)$ and the output of the query, i.e., $\mathcal{O}(\Elements, q)$. Hence, for an output generated by the LLM, we can calculate the output error $\epsilon_{\LLM}(\Elements, q)$.
Let $\Relevance_q(t_i)$ be either 0 or 1 for each token in $\Elements$\footnote{Check the experiments section (Section~\ref{sec:exp}) for practical examples.}.
Our goal is to estimate a set of unknown values $\{\Exposure_{\LLM}(i) \mid i \leq n\}$. 

After passing the task $(\Elements, q)$ to $\LLM$ with a specific ordering of $\Elements$, the error of the output provides aggregate information about the values $\Exposure_{\LLM}(i)$ based on the relevance scores. We model the relation between the error and the exposures as,

\begin{align}
    \frac{1}{\mathbb{E}[\epsilon_{\LLM}(\Elements, q)]} \propto \frac{1}{n} \sum^{n}_{i = 1} (\Exposure_{\LLM}(i) \cdot \Relevance_q(t_i))\label{eq:err-expo-rel}
\end{align}

In other words, the inverse of the error is directly proportional to the rank utilization of the items within \( \Elements \). The higher utilization implies that items of greater relevance are positioned in more exposed locations (\( \Exposure_{\LLM}(i) \)). Consequently, this results in a relatively smaller output error of the LLM.

\subsection{Estimation}

Let \(\pi_1, \pi_2, \dots, \pi_p\) be a set of \(p\) random permutations on \(\Elements\). 
We apply each permutation \(\pi_j\) on \(\Elements\) to obtain the permuted lists \(\Elements_{\pi_j}\). This would change the position of token $t_{\pi_j(i)}$ to $i$. Within each permuted list, the relevant tokens are positioned at random locations. Subsequently, we generate the output of the large language model (LLM) for each \(\Elements_{\pi_j}\) and compute the error \(\epsilon_{\LLM}(\Elements_{\pi_j}, q)\). This process allows us to sample exposures according to the relationship defined in equation~\eqref{eq:err-expo-rel}.

We then create an $n \times p$ matrix $\mathcal{R}$, such that $\mathcal{R}_{i, j} = \Relevance_q(t_{\pi_j(i)})$. Let $\vec{\epsilon}$ be a vector of size $p$ such that $\vec{\epsilon}_j = \frac{1}{\epsilon_{\LLM}(\Elements_{\pi_j}, q)}$. Now, we should solve the following equation to find the unknown exposure vector $\Exposure$:

\begin{align}
    \mathcal{R}^{\top} \cdot \Exposure = \vec{\epsilon}\label{eq:expo-matrix}
\end{align}

We can choose a value $p$ larger than $n$ to obtain a sufficient number of samples, thereby ensuring that the system of equations is overdetermined and can be solved effectively.

In order to solve equation ~\eqref{eq:expo-matrix}, we find an estimation $\overline{\Exposure}$ to minimize the Mean Squared Errors.

\begin{align*}
    MSE(\overline{\Exposure}) &= \|\mathcal{R}^{\top} \cdot \overline{\Exposure} - \vec{\epsilon}\|_2,\\
    \nabla MSE(\overline{\Exposure}) = 0 &\leftrightarrow \overline{\Exposure} = (\mathcal{R} \cdot \mathcal{R}^{\top})^{-1} \cdot \mathcal{R} \cdot \vec{\epsilon}
\end{align*}

As a result, $\overline{\Exposure}_i$ would be an estimation for $\Exposure_{\LLM}(i)$.

\subsection{Confidence}

While one can use a fixed budget on the number of permutations to use for estimating the exposure values, the user can alternatively specify a target variance in the estimation.
In such cases, considering the exponential search strategy, we start from a base number of permutations and compute the estimation variance, as explained in the following.
If the estimation variance is larger than the target variance, we double the number of permutations (i.e., double the value of $p$) and repeat the process.

We follow the standard confidence interval analysis for a Mean Squared Error estimation.
Each $\overline{\Exposure}_i$ is an unbiased estimator of $\Exposure_{\LLM}(i)$. This estimator follows a t-Distribution around the real value:

\begin{align}
    \frac{\overline{\Exposure}_i - \Exposure_{\LLM}(i)}{\sqrt{Var(\overline{\Exposure}_i)}} \sim t_{p - n}
\end{align}

Where $t_{p - n}$ is the t-Distribution with $p - n$ degree of freedom and,

\begin{align}
    Var(\overline{\Exposure}_i) = \hat{\sigma}^2 \cdot [(\mathcal{R} \cdot \mathcal{R}^{\top})^{-1}]_{i, i},\\
    \hat{\sigma} = \frac{\|\mathcal{R}^{\top} \cdot \overline{\Exposure} - \vec{\epsilon}\|_2}{p - n}
\end{align}

By increasing the number of random permutations $p$, we increase the degree of freedom and as a result we would have less variance and a narrower distribution for the estimation. 

\subsection{Practical Concerns}
The process for exposure value estimation requires $p$ separate API calls to the large language model, $\LLM$. In the following, we explore heuristic methods aimed at reducing the number of required API calls, thereby optimizing the associated costs.

At first, we can choose a task $(\Elements, q)$ such that there is only one token $t_i$ related to the query $q$. As a result, one can just place this relevant element into different positions and sample from each $\Exposure_{\LLM}(i)$ individually. Secondly, one can carefully create a task such that the result identifies how much information from all the positions is retained by the LLM. 

Specifically, consider the {\em token-counting task} that counts the number of occurrences of each token in the input $\Elements$.
For example, given the input $\Elements=\{a, b, b, b, a, a, a,c,c\}$, the output is {\tt ``a:4, b:3, c:2''}.
Note that in this example, all elements are relevant to the query, but each of them is relevant to one of the sub-problems (contributing only to one of the token counts).
Using this technique, each query to the LLM provides relevant information about {\em all} rank positions (not only a subset of positions that are relevant to the query).
We shall discuss this technique further with more examples in our experiments (Section~\ref{sec:exp}).

So far in this section, we examined the exposure at each token position within the input prompt. However, for specific tasks each element $e \in \Elements$ can correspond to multiple tokens.
Let $\ell(e)$ denote the number of tokens associated with element $e$. As outlined in Section~\ref{sec:sol}, the objective is to compute the utility of a particular ranking, which depends on the exposure of entire elements, rather than individual tokens, within the reranked input. To generalize the definition of utility, we define the exposure of each element $e$ as the average exposure of the tokens that compose it. In other words, we rewrite equation~\ref{eq:rerank2} as,

\[
\mathbb{E}[\text{utility}(\pi \mid q)] = 
\sum_{i=1}^{|\Elements|} 
\left( 
\mathbb{E} \left[ 
\frac{\sum_{j = i}^{i + \ell(e_i)} \Exposure_{\LLM}(i)}{\ell(e_i)}
\right] 
\cdot 
\mathbb{E} \left[ \Relevance_q(e_{\pi_i}) \right] 
\right)
\]
\section{Experiments}\label{sec:exp}

In this section, we assess the practical applicability of our results through experiments conducted on both real and synthetic datasets. Specifically, we evaluate two categories of tasks. The first category focuses on the Graph Degree Task, where a graph is constructed, and its edges are provided as input to the LLM, denoted as $\Elements$. A query $q$ is then issued, requesting the degree of a specific node within the graph. The second category involves answering queries about structured datasets. In these tasks, we supply a database table to the LLM ($\Elements$) and ask an aggregation query ($q$), expressed in natural language, about the table. We try different real-world datasets for this category. All experiments were conducted on a
local server equipped with 128 GB of memory, 32 CPU cores, and
two NVIDIA GeForce RTX 2080 Ti GPUs. The code is also accessible through \href{https://anonymous.4open.science/r/prompt-reranking-6638}{this repository}.

In the subsequent subsections, we begin by detailing the process of exposure discovery. Following this, we provide an in-depth discussion of the tasks, including the methods employed to construct and evaluate the LLMs. Next, we analyze the ranking performance of various open-source models (referred to as helpers) on these tasks. Finally, we compare the effectiveness of the different reranking techniques presented throughout this study.

\begin{figure}[!tb]
    \centering
    \begin{subfigure}[t]{0.36\textwidth}
        \includegraphics[width=\linewidth]{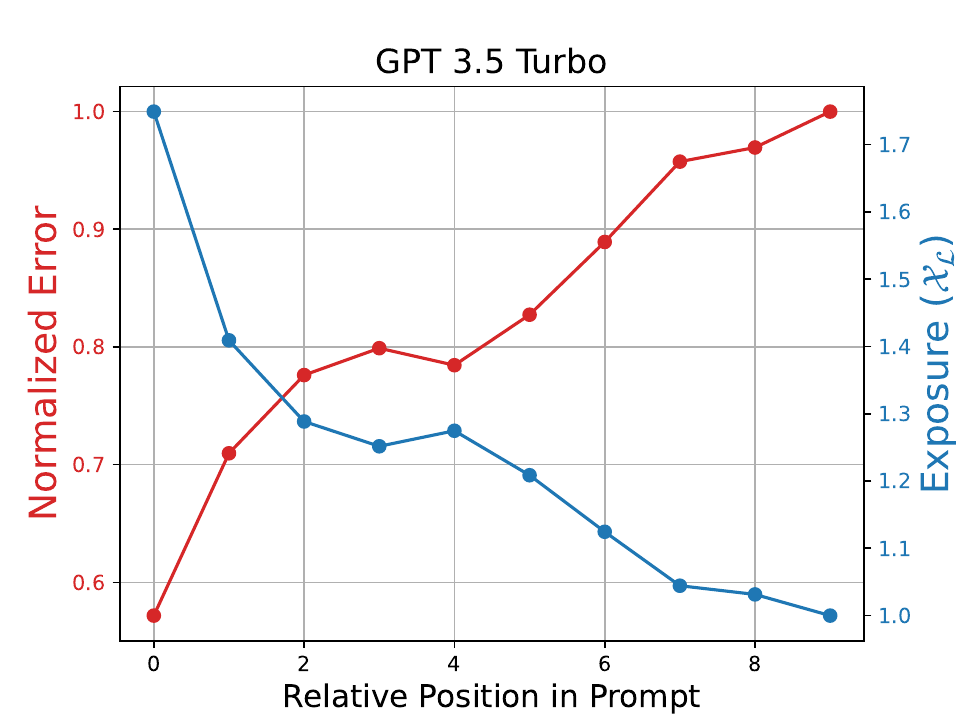}
        \caption{Relative exposure values on GPT 3.5 Turbo}
    \end{subfigure}
    \hfill
    \begin{subfigure}[t]{0.36\textwidth}
        \includegraphics[width=\linewidth]{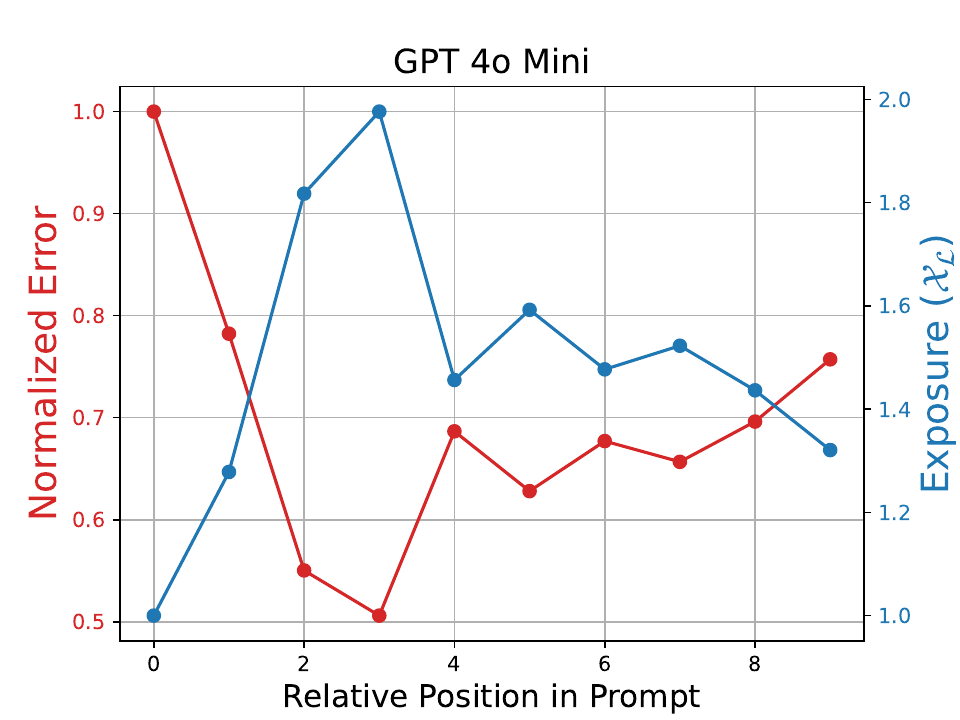}
        \caption{Relative exposure values on GPT 4o Mini}
    \end{subfigure}
    \caption{Exposure values for 'GPT-3.5 Turbo' and 'GPT-4o Mini.' The red plot represents the normalized error observed when placing relevant data at specific positions within a prompt of length 1000, averaged over 100 runs. In our model, the inverse of the average error at each position is proportional to the exposure $\Exposure_{\LLM}(i)$. Higher error at a given location indicates lower exposure at that index.}
    \label{fig:exposures}
\end{figure}

\subsection{Exposure Discovery}
In the pre-processing phase, we determine the exposure of each input token position to the LLM denoted as $\Exposure_{\LLM}(i)$. For our experiments, we utilize two widely adopted LLMs: GPT-3.5 Turbo and GPT-4o Mini. To measure token exposure, as outlined in Section ~\ref{sec:exposure}, we construct a synthetic task denoted as $(\Elements, q)$, where each element within $\Elements$ corresponds to an individual token. The query $q$ prompts the LLM to report the frequency of occurrence for each token in the input set $\Elements$. The LLM's output is represented as a key-value list, where each key corresponds to a token, and the associated value indicates the number of times that token was identified within the input.

We systematically reposition various tokens, including their repeated occurrences across consecutive positions (aka windows) within the input $\Elements$, to assess the extent to which each position is exposed to the LLM. We estimate the exposure levels by conducting multiple sampling iterations, as illustrated in Figure ~\ref{fig:exposures}. The errors are computed as the absolute difference between the actual repeats of each token and the corresponding value predicted by the LLM, averaged across all unique tokens. The exposure is modeled as the inverse of error at each position.

Based on Figure ~\ref{fig:exposures}, in the case of GPT-3.5 Turbo, the model has a tendency to prioritize the initial portion of a lengthy prompt, while the focus decreases towards the end of the prompt, increasing the likelihood that these latter tokens may get forgotten. However, GPT-4o Mini demonstrates a different pattern of token retention. It tends to forget the tokens at the beginning of the prompt, while those positioned in the middle are more likely to be remembered by the model.

We leverage the exposures identified in this subsection as a pre-processing step to address the subsequent tasks discussed in the following subsection.

\subsection{Tasks Setup}
In this subsection, we first provide a detailed overview of the synthetic Graph Degree Task, followed by an examination of the Database Query Task applied to real-world datasets.

\begin{table*}[ht]
\centering
\caption{Ranking utility comparison across algorithms and helper models. Each point is an average of 10 runs. The percentage values are the proximity of the numbers compared to the upper and lower bounds. The \color{DarkGreen}{green} \color{black} (\color{red}{red}\color{black}) arrow indicates the closeness to the upper (lower) bounds.}
\label{tab:rank-util}
\begin{tabular}{@{}lllllll@{}}
\toprule
\textbf{Algorithm} & \multicolumn{5}{c}{\textbf{Synthetic Graph Task}} \\ 
\cmidrule(lr){2-6}
& \textbf{DeepSeek-Coder-V2 (16B)} 
& \textbf{Gemma2 (9B)} 
& \textbf{Llama3.1 (8B)} 
& \textbf{Mistral (7B)} 
& \textbf{Qwen2 (7B)} \\ 
\midrule
\textbf{Optimum (UB)} 
& 3.02 (100\%) \uparrowgreen 
& 2.98 (100\%) \uparrowgreen 
& 3.01 (100\%) \uparrowgreen 
& 3.00 (100\%) \uparrowgreen 
& 2.98 (100\%) \uparrowgreen \\ \cmidrule[0.1mm](lr){1-6}
\textbf{Bipartite}         
& {\bf 2.95 (97\%)} \uparrowgreen 
& 1.87 (58\%) 
& {\bf 2.22 (70\%)} \uparrowgreen
& {\bf 2.49 (81\%)} \uparrowgreen 
& {\bf 1.03 (26\%)} \downarrowred \\
\textbf{Warm-up}             
& 0.67 (13\%) \downarrowred 
& {\bf 2.58 (85\%)} \uparrowgreen 
& 1.70 (51\%) 
& 2.03 (63\%) \uparrowgreen 
& 0.72 (15\%) \downarrowred \\ \cmidrule[0.1mm](lr){1-6}
\textbf{Random (LB)} 
& 0.31 (0\%) \downarrowred 
& 0.31 (0\%) \downarrowred 
& 0.32 (0\%) \downarrowred 
& 0.33 (0\%) \downarrowred 
& 0.32 (0\%) \downarrowred \\ 
\midrule
\addlinespace
\textbf{} & \multicolumn{5}{c}{\textbf{IMDB Dataset}} \\ 
\cmidrule(lr){2-6}
& \textbf{DeepSeek-Coder-V2 (16B)} 
& \textbf{Gemma2 (9B)} 
& \textbf{Llama3.1 (8B)} 
& \textbf{Mistral (7B)} 
& \textbf{Qwen2 (7B)} \\ 
\midrule
\textbf{Optimum (UB)} 
& 2.76 (100\%) \uparrowgreen 
& 2.60 (100\%) \uparrowgreen 
& 2.69 (100\%) \uparrowgreen 
& 2.52 (100\%) \uparrowgreen 
& 2.67 (100\%) \uparrowgreen \\ \cmidrule[0.1mm](lr){1-6}

\textbf{Bipartite}         
& {\bf 2.63 (94\%)} \uparrowgreen 
& 2.50 (95\%) \uparrowgreen 
& {\bf 2.48 (90\%)} \uparrowgreen 
& {\bf 2.22 (84\%)} \uparrowgreen 
& {\bf 1.60 (48\%)} \\

\textbf{Warm-up}           
& 1.30 (33\%) 
& {\bf 2.58 (99\%)} \uparrowgreen 
& 1.68 (52\%) 
& {\bf 2.22 (84\%)} \uparrowgreen 
& 1.50 (44\%) \\ \cmidrule[0.1mm](lr){1-6}

\textbf{Random (LB)} 
& 0.57 (0\%) \downarrowred 
& 0.48 (0\%) \downarrowred 
& 0.58 (0\%) \downarrowred 
& 0.55 (0\%) \downarrowred 
& 0.58 (0\%) \downarrowred \\
\midrule

\addlinespace
\textbf{} & \multicolumn{5}{c}{\textbf{OULAD Dataset}} \\ 
\cmidrule(lr){2-6}
& \textbf{DeepSeek-Coder-V2 (16B)} 
& \textbf{Gemma2 (9B)} 
& \textbf{Llama3.1 (8B)} 
& \textbf{Mistral (7B)} 
& \textbf{Qwen2 (7B)} \\ 
\midrule
\textbf{Optimum (UB)} 
& 2.78 (100\%) \uparrowgreen 
& 2.74 (100\%) \uparrowgreen 
& 2.78 (100\%) \uparrowgreen 
& 2.94 (100\%) \uparrowgreen 
& 2.83 (100\%) \uparrowgreen \\ \cmidrule[0.1mm](lr){1-6}

\textbf{Bipartite}         
& {\bf 2.76 (99\%)} \uparrowgreen 
& {\bf 2.73 (99\%)} \uparrowgreen 
& {\bf 2.67 (95\%)} \uparrowgreen 
& 2.73 (92\%) \uparrowgreen 
& {\bf 1.50 (45\%)} \\

\textbf{Warm-up}           
& 0.99 (27\%) \downarrowred 
& 0.63 (10\%) \downarrowred 
& 1.10 (32\%) 
& {\bf 2.90 (98\%)} \uparrowgreen 
& 1.44 (43\%) \\ \cmidrule[0.1mm](lr){1-6}

\textbf{Random (LB)} 
& 0.31 (0\%) \downarrowred 
& 0.38 (0\%) \downarrowred 
& 0.30 (0\%) \downarrowred 
& 0.35 (0\%) \downarrowred 
& 0.37 (0\%) \downarrowred \\
\bottomrule

\addlinespace
\textbf{} & \multicolumn{5}{c}{\textbf{Adults Dataset}} \\ 
\cmidrule(lr){2-6}
& \textbf{DeepSeek-Coder-V2 (16B)} 
& \textbf{Gemma2 (9B)} 
& \textbf{Llama3.1 (8B)} 
& \textbf{Mistral (7B)} 
& \textbf{Qwen2 (7B)} \\ 
\midrule
\textbf{Optimum (UB)} 
& 1.01 (100\%) \uparrowgreen 
& 1.53 (100\%) \uparrowgreen 
& 1.04 (100\%) \uparrowgreen 
& 1.60 (100\%) \uparrowgreen 
& 1.24 (100\%) \uparrowgreen \\ \cmidrule[0.1mm](lr){1-6}
\textbf{Bipartite}         
& {\bf 0.99 (97\%)} \uparrowgreen 
& {\bf 1.46 (95\%)} \uparrowgreen 
& {\bf 1.03 (99\%)} \uparrowgreen 
& 1.50 (92\%) \uparrowgreen 
& {\bf 0.72 (54\%)} \\
\textbf{Warm-up}           
& 0.39 (30\%) 
& 1.39 (90\%) \uparrowgreen 
& 0.26 (14\%) \downarrowred 
& {\bf 1.57 (98\%)} \uparrowgreen 
& 0.59 (42\%) \\ \cmidrule[0.1mm](lr){1-6}
\textbf{Random (LB)} 
& 0.12 (0\%) \downarrowred 
& 0.13 (0\%) \downarrowred 
& 0.13 (0\%) \downarrowred 
& 0.19 (0\%) \downarrowred 
& 0.11 (0\%) \downarrowred \\
\bottomrule
\end{tabular}
\end{table*}

\subsubsection{Graph Degree Task}
In this task, we generate a random graph using the Erdős-Rényi model ~\cite{erdos1960evolution}, where the edges represent the set of elements, denoted as $\Elements$. The primary query $q$ for this graph is: “What is the degree of a given node $v$?”. We systematically select various nodes $v$ within the graph and evaluate the responses provided by an LLM to these queries. Since the generated graph is fully accessible, we can compute the ground-truth degree of each queried node. To assess the accuracy of the LLM's responses, we calculate the error $\eps_{\LLM}$ as the absolute difference between the degree reported by the LLM and the ground-truth degree. 

We first pass the set of edges to the LLM and then ask the query in a different prompt message but in the same context. We selected graph sizes up to 500 edges to generate synthetic graphs to ensure that all edges could be represented within a single prompt when inputting them into LLMs, thereby adhering to the models' token limits.

\subsubsection{Database Query Task}
The second category of tasks involves providing a dataset table as input to the LLM and asking aggregation queries in natural language. For instance, a query might ask, "How many rows contain the value in the 'col' column equal to 'value'?"

We utilize three real-world datasets for this study. The IMDB Movies Dataset\cite{maas2011learning} provides information on the top 1,000 movies listed on IMDB. We extract a subset containing 60 rows from this dataset and formulate a query regarding the number of movies with a rating of $Rating \geq 8.2$. The Adult Income Dataset\cite{adult_2}, which comprises data on approximately 48,000 individuals for income prediction tasks, is also employed. From this dataset, we sample 60 rows and pose a query about the number of individuals associated with a specific 'workclass' category. Finally, the Open University Learning Analytics Dataset~\cite{kuzilek2017open} includes data on student enrollments across various courses. It contains 32K rows. For our task, we sample a subset of 100 rows and ask about the number of students enrolled in a particular course. These sample sizes are carefully selected to align with the token limits of LLMs.

\subsection{Analysis of the Ranking utility}
In this subsection, we utilize different open-source LLMs as helpers ($\Helper$) to estimate the relevance scores of elements to the query ($\Relevance_q$). We use five different models DeepSeek-Coder-v2 (16B) ~\cite{zhu2024deepseek}, Gemma2 (9B) ~\cite{team2024gemma}, Llama3.1 (8B) ~\cite{dubey2024llama}, Qwen2 (7B) ~\cite{yang2024qwen2}, and Mistral (7B) ~\cite{jiang2023mistral}.

We compare four ranking methods in this experiment. The first method, {\bf Warm-up}, as described in Section~\ref{sec:warm-up}, involves querying each partition of the input and subsequently splitting it. The second method, {\bf Bipartite}, detailed in Section~\ref{sec:bipartite}, utilizes Bipartite Graph Modeling to estimate relevance scores. To establish a lower bound for performance comparison, we use the {\bf Random} method, which involves randomly shuffling all input elements within the set $\Elements$ and passing them to the language model, $\LLM$. As an upper bound, we employ the {\bf Optimum} method, where elements are pre-sorted based on explicit knowledge of the task to maximize relevance to the query. This approach provides a theoretical upper limit for ranking utility.

Table ~\ref{tab:rank-util} presents the comparison of ranking utilities across different models and methods. The Random and Optimum methods serve as lower and upper bounds, respectively, providing benchmarks for comparison independent of any specific helper model. Due to the variability introduced by sampling, slight differences in the ranking utilities across different models can occur for these two methods, even though they are not using any helper models. Each value reported in the table represents the average of 10 independent runs.

The values represent the ranking utilities computed under the assumption that the exposure function is given by \(\Exposure_{\LLM}(i) = \frac{1}{i}\). In other words, once each helper model assigns relevance scores, a corresponding reranking function is obtained based on the specific model and method applied. To evaluate and compare the performance of different models, we analyze their respective rerankings. Specifically, we first sort the elements according to the reranking produced by each model. Within the sorted list, we then compute the utility associated with the truly relevant elements using exposure $\frac{1}{i}$. A higher utility value indicates that the relevant elements are positioned at higher ranks, reflecting that the reranking generated by the model assigns them higher estimated relevance scores.

The percentage values in the table indicate the {\bf Proximity} of the results, which can be defined as below.

\begin{definition}[Proximity]
Let \( L \) be the lower bound and \( U \) be the upper bound for a given observed error $x$. 
The \emph{proximity} \( P(x) \) with respect to these bounds can be defined as:
\[
P(x) = \frac{x - L}{U - L}, \quad \text{where} \quad L \leq x \leq U.
\]
\end{definition}

Based on this observation, in most cases, the {\bf Bipartite} approach estimates relevance scores nearly equivalent to those of the optimal solution. However, for certain models, such as Gemma2, the {\bf Warm-up} algorithm yields better results. This is because the Bipartite method requires the model to generate a list of scores, and the resulting output from the model may not always align with expectations. In contrast, for models designed to perform well in coding tasks, such as DeepSeek-Coder-v2 and Llama3.1, the Bipartite graph approach effectively enhances the reranking of the prompt.

\subsection{Analysis of the Output Error}
In this subsection, we compare the helper models and the methods in achieving the final goal which is enhancing the final output error from the LLM $\LLM$. The results for GPT-3.5 Turbo and GPT-4o Mini are presented in Tables ~\ref{tab:err-gpt3} and ~\ref{tab:err-gpt4}. For each helper model we used shorter names: DC2 (DeepSeek-Coder-v2), G2 (Gemma2), L3.1 (Llama3.1), M (Mistral), and Q2 (Qwen2). In these tables, Optimum is a lower bound since it is the best result one can achieve.

Each value in these tables represents the average of 10 runs. The errors for each helper model on a given dataset are normalized to the range \([0, 1]\). In most cases, the final error for the model employing the Bipartite method is close to the optimal solution. However, certain helper models, such as Qwen2, perform poorly in reranking the prompt, resulting in errors comparable to those obtained from random shuffling of the input. We observe that the Adults dataset presents a relatively more challenging task in most cases. This increased difficulty arises from the larger number of columns in this dataset compared to others, with a significant portion of these columns consisting of string-type data. The prevalence of such features complicates the query-answering process for the LLM.

\begin{table*}[!tbh]
    \centering
    \caption{Normalized Error ($\eps_{\LLM}$) across algorithms and helper models. The errors are normalized for each helper model to align them in the interval $[0, 1]$. Each value is an average of 10 runs. The \color{DarkGreen}{green} \color{black} (\color{red}{red}\color{black}) arrow indicates the closeness to the lower (upper) bound.}
    \begin{subtable}{0.45\textwidth}
        \centering
        \caption{Output error results on GPT-4o Mini}
        
\begin{tabular}{@{}lllllll@{}}
\toprule
\textbf{Algorithm} & \multicolumn{5}{c}{\textbf{Synthetic Graph Task}} \\ 
\cmidrule(lr){2-6}
& \textbf{DC2} 
& \textbf{G2} 
& \textbf{L3.1} 
& \textbf{M} 
& \textbf{Q2} \\
\midrule
\textbf{Random (UB)} 
& 1.00 \uparrowred 
& 1.00 \uparrowred 
& 1.00 \uparrowred 
& 1.00 \uparrowred 
& 1.00 \uparrowred \\ \cmidrule[0.1mm](lr){1-6}
\textbf{Warm-up}             
& 0.98 \uparrowred
& {\bf 0.12} \downarrowgreen
& 0.99 \uparrowred
& 0.5
& 0.72 \uparrowred \\ 
\textbf{Bipartite}         
& {\bf 0.12} \downarrowgreen
& 0.63
& {\bf 0.68}
& {\bf 0.12} \downarrowgreen
& {\bf 0.28} \downarrowgreen \\ \cmidrule[0.1mm](lr){1-6}
\textbf{Optimum (LB)} 
& 0.00 \downarrowgreen 
& 0.00 \downarrowgreen 
& 0.00 \downarrowgreen 
& 0.00 \downarrowgreen  
& 0.00 \downarrowgreen \\ 
\midrule
\addlinespace
\textbf{} & \multicolumn{5}{c}{\textbf{IMDB Dataset}} \\ 
\cmidrule(lr){2-6}
& \textbf{DC2} 
& \textbf{G2} 
& \textbf{L3.1} 
& \textbf{M} 
& \textbf{Q2} \\ 
\midrule
\textbf{Random (UB)} 
& 1.00 \uparrowred 
& 1.00 \uparrowred 
& 1.00 \uparrowred 
& 1.00 \uparrowred 
& 1.00 \uparrowred \\ \cmidrule[0.1mm](lr){1-6}

\textbf{Warm-up}           
& 0.33
& {\bf 0.11} \downarrowgreen
& 0.64
& {\bf 0.42}
& 0.68 \\ 

\textbf{Bipartite}         
& {\bf 0.25} \downarrowgreen
& 0.10 \downarrowgreen
& {\bf 0.01} \downarrowgreen
& {\bf 0.42}
& {\bf 0.62} \\ \cmidrule[0.1mm](lr){1-6}

\textbf{Optimum (LB)} 
& 0.00 \downarrowgreen 
& 0.00 \downarrowgreen 
& 0.00 \downarrowgreen 
& 0.00 \downarrowgreen  
& 0.00 \downarrowgreen \\
\midrule

\addlinespace
\textbf{} & \multicolumn{5}{c}{\textbf{OULAD Dataset}} \\ 
\cmidrule(lr){2-6}
& \textbf{DC2} 
& \textbf{G2} 
& \textbf{L3.1} 
& \textbf{M} 
& \textbf{Q2} \\
\midrule
\textbf{Random (UB)} 
& 1.00 \uparrowred 
& 1.00 \uparrowred 
& 1.00 \uparrowred 
& 1.00 \uparrowred 
& 1.00 \uparrowred \\ \cmidrule[0.1mm](lr){1-6}

\textbf{Warm-up}           
& 0.17 \downarrowgreen 
& 0.94 \uparrowred
& 0.57
& {\bf 0.68}
& 0.93 \uparrowred \\ 

\textbf{Bipartite}         
& {\bf 0.02} \downarrowgreen
& {\bf 0.01} \downarrowgreen
& {\bf 0.28} \downarrowgreen
& 0.81 \uparrowred
& {\bf 0.86} \uparrowred \\ \cmidrule[0.1mm](lr){1-6}

\textbf{Optimum (LB)} 
& 0.00 \downarrowgreen 
& 0.00 \downarrowgreen 
& 0.00 \downarrowgreen 
& 0.00 \downarrowgreen  
& 0.00 \downarrowgreen \\
\bottomrule

\addlinespace
\textbf{} & \multicolumn{5}{c}{\textbf{Adults Dataset}} \\ 
\cmidrule(lr){2-6}
& \textbf{DC2} 
& \textbf{G2} 
& \textbf{L3.1} 
& \textbf{M} 
& \textbf{Q2} \\ 
\midrule

\textbf{Random (UB)} 
& 1.00 \uparrowred 
& 1.00 \uparrowred 
& 1.00 \uparrowred 
& 1.00 \uparrowred 
& 1.00 \uparrowred \\ \cmidrule[0.1mm](lr){1-6}

\textbf{Warm-up}           
& 1.00 \uparrowred
& 0.31
& 0.84 \uparrowred
& {\bf 0.22} \downarrowgreen
& {\bf 0.71} \\ 

\textbf{Bipartite}         
& {\bf 0.57}
& {\bf 0.28} \downarrowgreen
& {\bf 0.22} \downarrowgreen
& 0.23 \downarrowgreen
& {\bf 0.71} \\ \cmidrule[0.1mm](lr){1-6}

\textbf{Optimum (LB)} 
& 0.00 \downarrowgreen 
& 0.00 \downarrowgreen 
& 0.00 \downarrowgreen 
& 0.00 \downarrowgreen  
& 0.00 \downarrowgreen \\
\bottomrule
\label{tab:err-gpt4}
\end{tabular}

    \end{subtable}%
    \hfill
    \begin{subtable}{0.45\textwidth}
        \centering
        \caption{Output error results on GPT-3.5 Turbo}
        
\begin{tabular}{@{}lcccccc@{}}
\toprule
\textbf{Algorithm} & \multicolumn{5}{c}{\textbf{Synthetic Graph Task}} \\ 
\cmidrule(lr){2-6}
& \textbf{DC2} 
& \textbf{G2} 
& \textbf{L3.1} 
& \textbf{M} 
& \textbf{Q2} \\ 
\midrule
\textbf{Random (UB)} 
& 1.00 \uparrowred 
& 1.00 \uparrowred 
& 1.00 \uparrowred 
& 1.00 \uparrowred 
& 1.00 \uparrowred \\ \cmidrule[0.1mm](lr){1-6}
\textbf{Warm-up}             
& 0.72
& {\bf 0.35}
& 0.63
& 0.90 \uparrowred
& 0.58 \\
\textbf{Bipartite}         
& {\bf 0.09} \downarrowgreen
& 0.59
& {\bf 0.37}
& {\bf 0.14} \downarrowgreen
& {\bf 0.30} \\ \cmidrule[0.1mm](lr){1-6}
\textbf{Optimum (LB)} 
& 0.00 \downarrowgreen 
& 0.00 \downarrowgreen 
& 0.00 \downarrowgreen 
& 0.00 \downarrowgreen  
& 0.00 \downarrowgreen \\ 
\midrule
\addlinespace
\textbf{} & \multicolumn{5}{c}{\textbf{IMDB Dataset}} \\ 
\cmidrule(lr){2-6}
& \textbf{DC2} 
& \textbf{G2} 
& \textbf{L3.1} 
& \textbf{M} 
& \textbf{Q2} \\ 
\midrule
\textbf{Random (UB)} 
& 1.00 \uparrowred 
& 1.00 \uparrowred 
& 1.00 \uparrowred 
& 1.00 \uparrowred 
& 1.00 \uparrowred \\ \cmidrule[0.1mm](lr){1-6}

\textbf{Warm-up}           
& 0.56
& 0.87 \uparrowred
& 0.85 \uparrowred
& {\bf 0.49}
& 0.50 \\ 

\textbf{Bipartite}         
& {\bf 0.03} \downarrowgreen
& {\bf 0.29} \downarrowgreen
& {\bf 0.04} \downarrowgreen
& 0.69
& {\bf 0.48} \\ \cmidrule[0.1mm](lr){1-6}

\textbf{Optimum (LB)} 
& 0.00 \downarrowgreen 
& 0.00 \downarrowgreen 
& 0.00 \downarrowgreen 
& 0.00 \downarrowgreen  
& 0.00 \downarrowgreen \\
\midrule

\addlinespace
\textbf{} & \multicolumn{5}{c}{\textbf{OULAD Dataset}} \\ 
\cmidrule(lr){2-6}
& \textbf{DC2} 
& \textbf{G2} 
& \textbf{L3.1} 
& \textbf{M} 
& \textbf{Q2} \\ 
\midrule
\textbf{Random (UB)} 
& 1.00 \uparrowred 
& 1.00 \uparrowred 
& 1.00 \uparrowred 
& 1.00 \uparrowred 
& 1.00 \uparrowred \\ \cmidrule[0.1mm](lr){1-6}

\textbf{Warm-up}           
& 0.42
& 0.71
& 0.79 \uparrowred
& 0.55
& 0.84 \uparrowred \\ 

\textbf{Bipartite}         
& {\bf 0.11} \downarrowgreen
& {\bf 0.04} \downarrowgreen
& {\bf 0.64}
& {\bf 0.32}
& {\bf 0.75} \uparrowred \\ \cmidrule[0.1mm](lr){1-6}

\textbf{Optimum (LB)} 
& 0.00 \downarrowgreen 
& 0.00 \downarrowgreen 
& 0.00 \downarrowgreen 
& 0.00 \downarrowgreen  
& 0.00 \downarrowgreen \\
\bottomrule

\addlinespace
\textbf{} & \multicolumn{5}{c}{\textbf{Adults Dataset}} \\ 
\cmidrule(lr){2-6}
& \textbf{DC2} 
& \textbf{G2} 
& \textbf{L3.1} 
& \textbf{M} 
& \textbf{Q2} \\ 
\midrule

\textbf{Random (UB)} 
& 1.00 \uparrowred 
& 1.00 \uparrowred 
& 1.00 \uparrowred 
& 1.00 \uparrowred 
& 1.00 \uparrowred \\ \cmidrule[0.1mm](lr){1-6}

\textbf{Warm-up}           
& 0.62
& 0.85 \uparrowred
& 0.55
& {\bf 0.50}
& 0.71 \uparrowred \\ 

\textbf{Bipartite}         
& {\bf 0.21} \downarrowgreen
& {\bf 0.71} \uparrowred
& {\bf 0.11} \downarrowgreen
& 0.62
& {\bf 0.42} \\ \cmidrule[0.1mm](lr){1-6}

\textbf{Optimum (LB)} 
& 0.00 \downarrowgreen 
& 0.00 \downarrowgreen 
& 0.00 \downarrowgreen 
& 0.00 \downarrowgreen  
& 0.00 \downarrowgreen \\
\bottomrule
\label{tab:err-gpt3}
\end{tabular}

    \end{subtable}
    \label{tab:errs-all}
\end{table*}

\subsection{The Effect of Exposure Discovery}
\begin{figure}[!tb]
    \centering
    \begin{subfigure}[b]{0.36\textwidth}
        \centering
        \includegraphics[width=\textwidth]{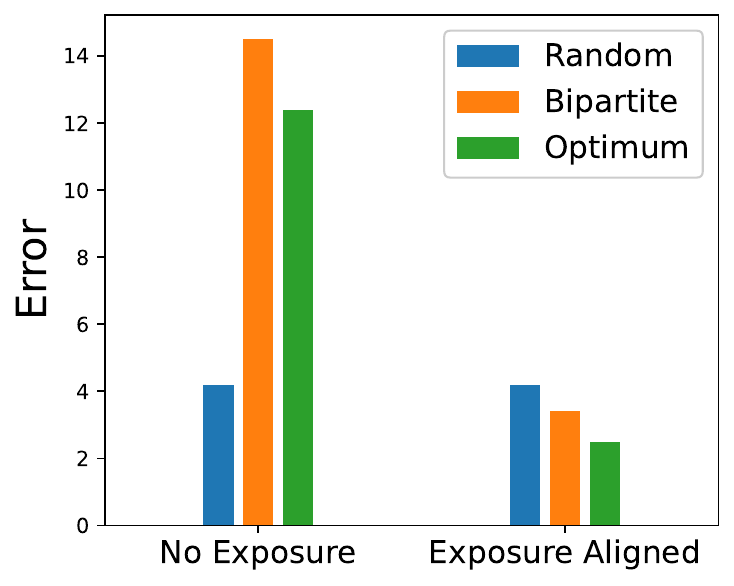}
        \caption{IMDB Dataset}
        \label{fig:subfig1}
    \end{subfigure}

    \begin{subfigure}[b]{0.36\textwidth}
        \centering
        \includegraphics[width=\textwidth]{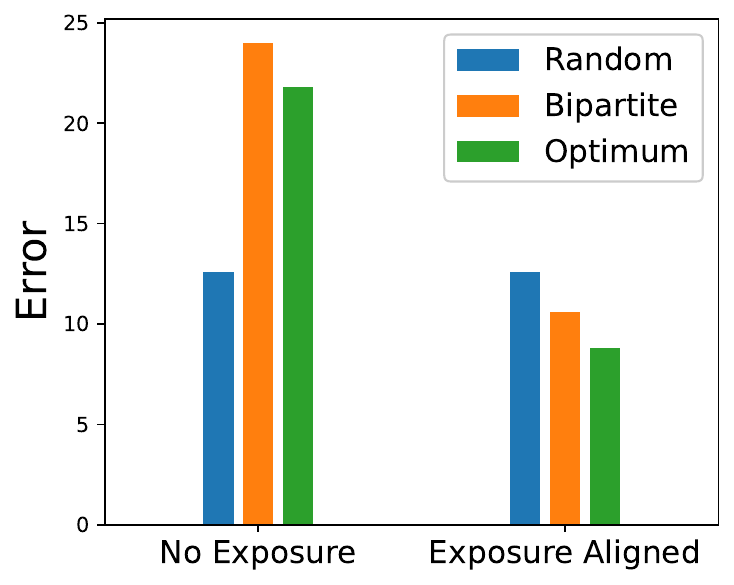}
        \caption{OULAD Dataset}
        \label{fig:subfig2}
    \end{subfigure}
    \caption{Verifying the effect of the exposure function $\Exposure_{\LLM}$ on the sorted list for GPT-4o Mini. For this LLM, sorting $\Elements$ in descending order results in the highest error rate. However, applying the exposure function significantly reduces the error. The helper LLM for this result is DeepSeek-Coder-v2.}
    \label{fig:exposure_effect}
\end{figure}

In this subsection, we evaluate the impact of exposure discovery on the final outcomes of the proposed methods. As illustrated in Figure~\ref{fig:exposures}, GPT-4o Mini exhibits an unexpected behavior by focusing more on the middle portion of the prompt rather than the beginning. In Figure~\ref{fig:exposure_effect}, we present a comparative analysis of two scenarios. In the first scenario, exposures are not utilized, and the input prompts are sorted in descending order based solely on the estimated relevance scores obtained through various methods. In the second scenario, exposures are incorporated, as previously described, to refine the reranking of the input prompts.

The results demonstrate that, for both the IMDB and OULAD datasets, applying the exposure significantly reduces the final error of the GPT-4o model. This finding indicates that the insights gained from the exposure discovery process as a pre-processing step are effective in generalizing various tasks during query time. Moreover, it confirms that this pre-processing approach serves as a valuable step in addressing the underlying problem.

\section{Related Work}\label{sec:related}

In this section, we review the literature relevant to our research. The section is organized into distinct categories, discussed in detail within the subsequent subsections.

\subsection{Handling Long Inputs}
The challenge of handling long input prompts is a widely recognized issue in large language models. Various studies have approached this problem from multiple perspectives. One notable scenario occurs when prompts involve in-context learning ~\cite{brown2020language,xie2021explanation}, which often results in extended inputs that LLMs struggle to fully process. The "Lost in the Middle" phenomenon, as identified by Liu et al.~\cite{liu2024lost}, highlights this issue, where LLMs fail to retain or utilize certain portions of lengthy inputs. Other studies have approached this issue by modifying the training datasets used for LLMs~\cite{he2024never}.

Several studies, have explored modifying the architecture of LLMs, particularly the attention mechanisms, to improve their ability to process extensive context and handle larger input sizes \cite{chen2023extending, li2020linear, child2019generating, zaheer2020big, beltagy2020longformer, hsieh2024found}. 

The study by Li et al.~\cite{li2024long} assesses the ability of LLMs to perform in-context learning (ICL) with extended input sequences. The findings underscore the challenges LLMs encounter when attempting to scale in-context learning to longer sequences.

Chen et al.~\cite{chen2023walking} explore an approach for handling long contexts by iteratively interacting with LLMs. In their method, a tree of summary nodes is first generated from the input prompt, and upon receiving a query, the system searches within the tree to retrieve relevant information. This technique addresses the challenge of processing long contexts by structuring and segmenting information for more efficient retrieval.

\subsection{Prompt Compression}
Another line of research focuses on compressing long input prompts while preserving sufficient information for the LLM to generate accurate responses ~\cite{li2023unlocking, jiang2023llmlingua, jiang2023longllmlingua}. This approach differs fundamentally from our problem, as our objective is to preserve all information from the prompt without any loss. Additionally, we focus on symmetric tasks where the ordering of inputs should not affect the final outcome.

The "Selective Context" method proposed by Li et al.~\cite{li2023unlocking} aims to filter out irrelevant portions of the input prompt by estimating the self-information of different segments, such as sentences or tokens, to determine which parts are most important for the LLM's response. A smaller base language model is employed for this estimation. However, their approach assumes access to the output probabilities of the model, whereas our method operates under the assumption of black-box access to any LLM.

The other approaches, such as LLMLingua ~\cite{jiang2023llmlingua}, attempt to compress the input to large language models (LLMs) by using a base LLM to identify relevant information within the prompt. This method assumes conditional dependencies between tokens in the prompt and seeks to estimate the associated probabilities using the base LLM. Additionally, the approach leverages the output probabilities from the LLM to refine the relevance detection process.

Machlab et al.~\cite{machlab2024llm} investigate the recall patterns of LLMs in relation to input prompts, focusing on how recall is influenced by both the length of the prompt and the position of relevant information (referred to as the "needle in a haystack" problem). Their findings indicate that recall patterns are heavily dependent on the structure and content of the prompt. This bears some similarity to our approach for exposure discovery, though we focus on a specific set of tasks, known as symmetric problems, to estimate these exposures. Our experiments demonstrate that this recall pattern is consistent across different tasks within the same category.

\subsection{LLMs for Ranking}
The use of LLMs for ranking a set of objects has been extensively studied in the literature. Several works focus on leveraging LLMs to rank sets of documents or passages~\cite{zhuang2024setwise, sachan2022improving, liang2022holistic, nogueira2020document, zhuang2021deep}. Additionally, other studies address document ranking as a subtask within the broader framework of Retrieval-Augmented Generation (RAG)~\cite{zhuang2021tilde, yu2024rankrag}.

Zhuang et al.~\cite{zhuang2024setwise} discuss various strategies for ranking with LLMs, including setwise, pointwise, and pairwise approaches. Their findings show that the pairwise method is the most effective, while the pointwise approach is the most efficient. However, the pointwise method presents challenges due to instability caused by biases and the inherent non-deterministic behavior of LLMs. In this work, we address this challenge by employing a bipartite graph approach to remove the biases in the pointwise approach.

\subsection{Peer-review Process}
The existing literature on the peer-review and peer-grading process is also related to our work. Several studies aim to address the challenges in these processes, such as mitigating biases, defining the incentives of peer reviewers, and eliminating adversarial behaviors~\cite{de2016incentives, jecmen2020mitigating, chakraborty2018removing, hamer2005method, cho2007scaffolded, shah2013case}. 

Chakraborty et al.~\cite{chakraborty2018removing} model peer grading as a game-theoretic problem, aiming to create incentives and establish equilibrium among peers. Their approach addresses strategic behavior in peer grading and incorporates mechanisms to incentivize honest assessments. The model is designed to be bias-insensitive, using a small set of probe papers to detect bias in individual reviews.

\section{Discussions}\label{sec:discussion}
In this section, we first examine the advantages of our method in practical applications, followed by a discussion of its limitations.

\subsection{Advantages}
In all the proposed models and methods, we have the assumption that the final large language model is treated as a black-box, meaning we have no explicit knowledge of its internal architecture and make no use of such details. Even the smaller helper models employed in our approach are more cost-effective language models, yet we similarly assume no access to or detailed understanding of their underlying structures. Consequently, our method can be regarded as a wrapper that complements any advancements in large language models and improvements in their accuracy. This approach is applicable to any given LLM to enhance its performance in addressing symmetric problems, functioning as an additional layer to increase overall accuracy. In general, without detailed knowledge of a model's architecture, it is challenging to understand how it remembers different parts of the input. However, our approach aims to estimate the exposure of different input segments, allowing us to infer the recall patterns of a black-box LLM.

Additionally, we adopt an abstract approach from the problem perspective. Without requiring explicit knowledge of the specific problem or the query applied to the bag of elements, we aim to identify a re-ranking function that optimizes the final accuracy. This approach serves as a generalizable solution applicable to any symmetric problem (a query asked about a set or multi-set of elements). 

The proposed model and algorithm for debiasing the LLM evaluations, known as bipartite evaluation, can be applied in any scenarios where pointwise evaluations are performed on a set of objects and varying biases exist across different evaluations. More broadly, the approach is applicable to any problem that can be framed as a peer-review (peer-grading) process.

\subsection{Limitations}
As discussed in previous sections, our focus is on symmetric problems, where a set (or multi-set) of elements $\Elements$ is presented, and a query is asked about it. 
Our methods aim to re-rank $\Elements$ under this assumption and, therefore, are not applicable in cases where the input is an ordered list, and the ordering of elements plays a crucial role in the final response to the query.

Additionally, 
we assume that different elements have different relevance when answering the given query.
Hence, placing the high-relevance elements in highly exposed positions within the input sequence leads to improved accuracy. Conversely, when all elements in the input set are nearly equally relevant to the query, re-ranking the input may result in only marginal accuracy improvements. Regardless of the sorting method, some highly relevant elements may still stay in less exposed locations, leading to potential degradation in the LLM's ability to retain them effectively.

Finally, in a practical setting, the proposed approaches necessitate the use of a significantly smaller helper model in conjunction with the primary large language model. As a result, deploying these methods requires the additional deployment of a smaller model. 
\section{Conclusion}\label{sec:conclusion}

In this work, we addressed the challenge of handling lengthy input prompts for large language models within the context of {\em symmetric tasks}. 
We observe that while the performance of LLMs drops for large inputs, certain positions in the input are less likely to be missed by an LLM.
Moreover, reordering the input elements of symmetric tasks does not logically affect the query outcome. 

Following these observations, we introduced the LLM input reranking problem to reorder an input in a way that the accuracy of the LLM is maximized for the given query.

We proposed a two-step solution for this problem. First, during the preprocessing phase, we identify the exposure of ranking positions for a given LLM. Next, at query time, we estimate the relevance score of each element to the query. By combining these insights, we rerank the input and pass the reranked data to the LLM. For query relevance estimation, we introduced a method based on a bipartite graph modeling, with a performance to the optimal reranking in our experiments. 

Our solutions treat both the tasks and the LLMs as black-box components, allowing for a high level of abstraction.
As a result, those can  serve as a wrapper layer on top of any of the existing or future LLMs for solving symmetric tasks.

Despite the contributions of this work, certain limitations remain. For instance, the deployment of smaller LLMs and their suboptimal performance in score estimation in certain cases present challenges. By addressing these limitations and exploring enhancements in the reranking approaches, future research has the potential to further advance the effectiveness and scalability of language models in real-world applications.


\bibliographystyle{ACM-Reference-Format}
\bibliography{ref}

\end{document}